\documentclass[runningheads]{llncs}
\usepackage[utf8]{inputenc}
\usepackage{graphicx}
\usepackage[table,xcdraw]{xcolor}
\usepackage{gensymb}
\usepackage{url}
\usepackage{color}
\usepackage[caption=false]{subfig}
\usepackage{float}
\usepackage{cite}
\usepackage{amssymb}
\usepackage{placeins}

\newcolumntype{L}[1]{>{\raggedright\let\newline\\\arraybackslash\hspace{0pt}}m{#1}}
\newcolumntype{C}[1]{>{\centering\let\newline\\\arraybackslash\hspace{0pt}}m{#1}}
\newcolumntype{R}[1]{>{\raggedleft\let\newline\\\arraybackslash\hspace{0pt}}m{#1}}
\begin{document}
\title{RobôCIn Small Size League\\ Extended Team Description Paper for RoboCup 2023}

%
%
\author{Aline Oliveira \and Cauê Gomes \and Cecília Silva \and Charles M. Alves \and Danilo Souza \and Driele Xavier \and Edgleyson Silva \and Felipe Martins \and Lucas H. Cavalcanti \and Lucas Maciel \and Matheus Paixão \and Matheus Vasconcelos \and Matheus Vinícius \and João G. Melo \and João P. Moura \and José R. Silva \and José V. Cruz \and Pedro H. Santana \and Pedro P. Oliveira \and Riei Rodrigues \and Roberto Fernandes \and Ryan Morais \and Tamara Teobaldo \and Washington Silva \and Edna Barros}
%
%
\institute{Centro de Informática, Universidade Federal de Pernambuco. \\
Av. Prof. Moraes Rego, 1235 - Cidade Universitária, Recife - Pernambuco, Brazil.
\email{robocin@cin.ufpe.br}\\
\url{https://robocin.com.br/}}
\maketitle              

\begin{abstract}
RobôCIn has participated in RoboCup Small Size League since 2019, won its first world title in 2022 (Division B), and is currently a three-times Latin-American champion. This paper presents our improvements to defend the Small Size League (SSL) division B title in RoboCup 2023 in Bordeaux, France. This paper aims to share some of the academic research that our team developed over the past year. Our team has successfully published 2 articles related to SSL at two high-impact conferences: the 25th RoboCup International Symposium and the 19th IEEE Latin American Robotics Symposium (LARS 2022). Over the last year, we have been continuously migrating from our past codebase to Unification. We will describe the new architecture implemented and some points of software and AI refactoring. In addition, we discuss the process of integrating machined components into the mechanical system, our development for participating in the vision blackout challenge last year and what we are preparing for this year.

\keywords{RobôCIn \and RoboCup 2023 \and Robotics \and Small Size League}
\end{abstract}

\section{Hardware}
 The hardware updates for this year aim at more reliable motion control by improving our mechanics project. We have added a brass thread to our aluminum drive transmission support, preventing it from wearing out due to the shaft's friction. Besides the hardware adaptations for the Vision Blackout challenge, which are detailed in Subsection \ref{sec:blackout-hardware}, our electronics project has remained unchanged, and general hardware specifications can be found in Table \ref{tab:specifications}, with no changes from the 2020 version.

\begin{table}[]
\centering
\caption{Robot Specifications}
\label{tab:specifications}
\def\arraystretch{1.1}
\resizebox{0.65\columnwidth}{!}{%
\begin{tabular}{|l|l|}
\hline
\textbf{Robot Version} & \textbf{v2022}         \\ \hline
Driving motors         & Maxon EC-45 flat - 50W \\ \hline
Max \% ball coverage   & 19.55\%                \\ \hline
Microcontroller        & STM32F767ZI            \\ \hline
Gear Transmission      & 18 : 60                \\ \hline
Gear Type              & External Spur          \\ \hline
Wheel                  & 3D Printed             \\ \hline
Total Weight           & 2.36 kg                \\ \hline
Dribbling motor        & Maxon EC-max 22, 25W   \\ \hline
Encoder                & \textit{MILE 1024 CPT} \\ \hline
Dribbling Gear         & 1 : 1 : 1              \\ \hline
Dribbling bar diameter & 13mm                   \\ \hline
Max. kick speed        & 6.5m/s                 \\ \hline
Communication Link     & nRF24L01+              \\ \hline
Battery                & LiPo 2200mah 4S 35C    \\ \hline
\end{tabular}%
}
\end{table}

\subsection{Drive Transmission Support} \label{dribbling}



\begin{figure}[ht]
	\centering
	\captionsetup[subfloat]{justification=centering}
    \subfloat[RoboCup support version.]{\includegraphics[width=.32\linewidth, scale=1]{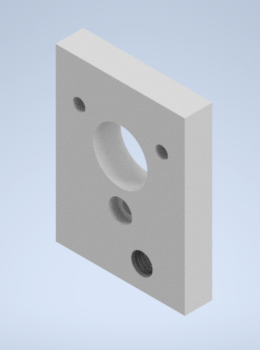}}
    \hspace{2cm}
    \subfloat[LARC support version with brass thread.]{\includegraphics[width=.32\linewidth, scale=1]{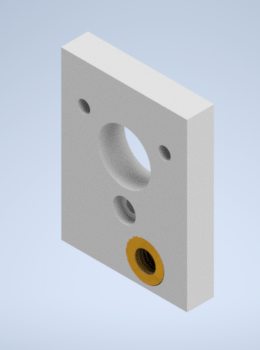}}
	\caption{Drive transmission support versions used in the RoboCup and LARC 2022.}
	\label{fig:driveTransmission}
\end{figure}

To achieve our goal of competing in Division A in the following years, we need to minimize errors in the robot's movements, and improving our drive set is a necessary step. For the mechanics project, our team uses mainly 3D-printed parts, however, this approach has shown to be insufficient for building a reliable transmission set due to resistance and precision limitations from the adopted materials (PLA) and the 3D printing process. Thus, for RoboCup 2022, we added a machined aluminum drive transmission support to our robots, Figure \ref{fig:driveTransmission}(a), making the transmission system smoother and ensuring the transmission gears' tolerance was respected.

Even though the machined transmission approach has improved the system, during RoboCup 2022, some of the drive sets presented looseness on the wheel shaft, and we partially solved the problem by grounding the wheel axis with Tekbond 793. Therefore, after the competition, we conducted a material analysis and found that the major reason for the gap was that the hardness of our stainless steel wheel shaft was much higher than the hardness of the aluminum drive transmission support, causing it to wear out due to the forces applied to this structure.

To mitigate this problem, we added a brass thread to the aluminum support, 
Figure \ref{fig:driveTransmission}(b), which has a greater hardness than the material from the previous model, \ref{fig:driveTransmission}(a), equalizing the contact forces. This model was used in the 2022 Latin American Robotics Competition (LARC) and has shown more reliability than the previous, full-aluminum, version.

We also had problems with tolerance in the machining process of the drive transmission support to add the brass thread. Figure \ref{fig:driveTransmission}(c) shows the consequences of this inaccuracy in the hole, causing backlash problems for the gearing misaligned axis. These problems were solved by manufacturing new parts of the transmission supports in collaboration with the Physics Department of our university, which supported us with high-precision machines that guarantee the reliability of our robots. 

\begin{figure}[ht]
\centering
\includegraphics[width=.44\textwidth]{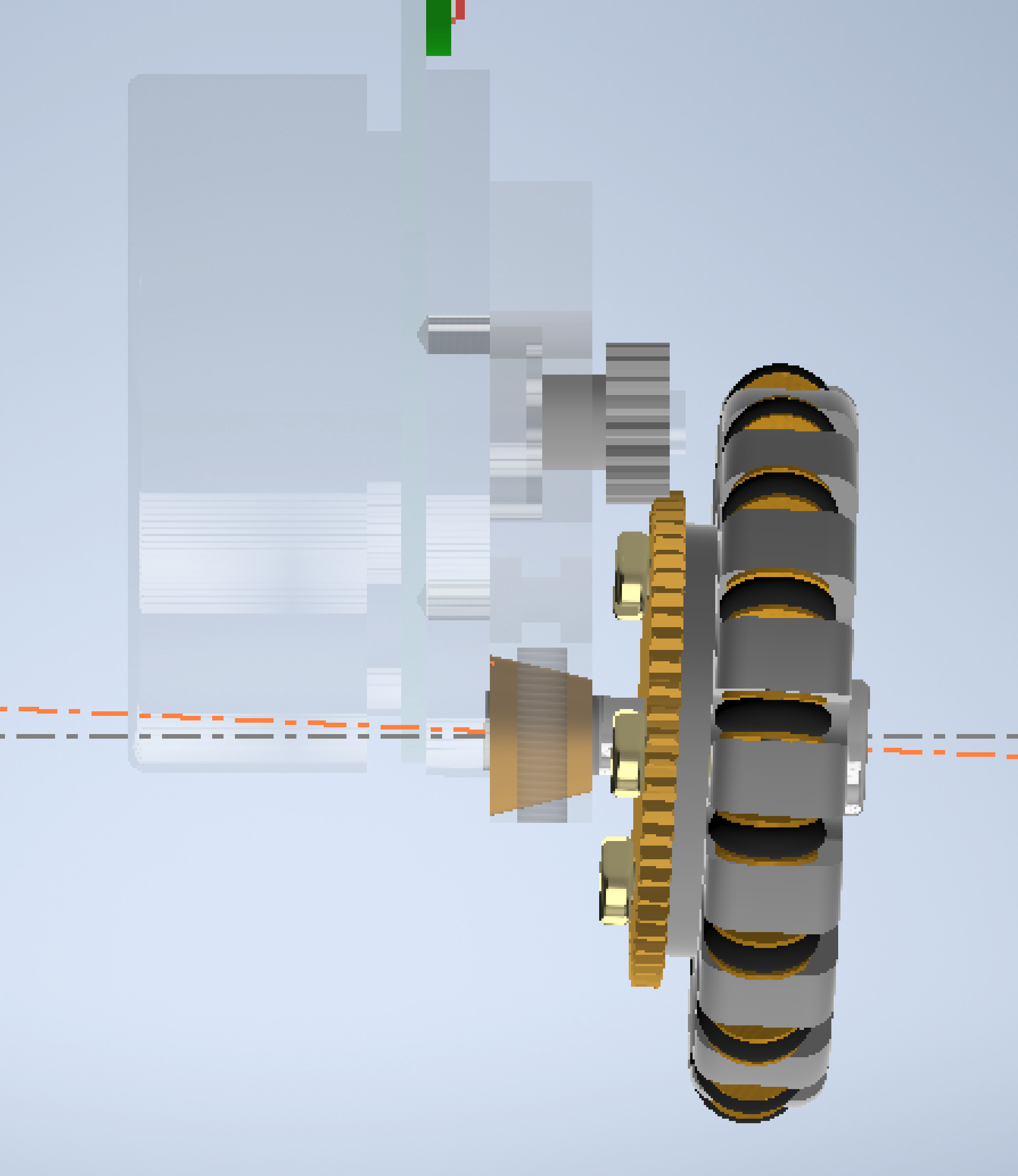}
\caption{Uncentric brass inserted thread.}
\label{fig:Brass insert nut}
\end{figure}




\section{Vision Blackout Challenge} \label{vision-blackout}

For participating in the Vision Blackout Challenge, hardware adaptations were made, new low-level navigation methods were implemented and a complete software infrastructure was built for allowing our robots to execute SSL soccer skills autonomously. Our goal was to create a robust enough infrastructure to implement each robot skills by only creating new Finite State Machines (FSM), all using onboard modules for sensing and processing. With this architecture, we were able to complete 2 of the 4 stages of the challenge in 2022's competition, achieving 2nd place. Also, we shared details of our research on recent papers \cite{towards-an-autonomous-ssl-robot, objects-detection-and-position-estimation, dataset-ssl} and open-source project datasets and documentation\footnote{https://github.com/bebetocf/ssl-dataset}\footnote{https://github.com/jgocm/ssl-detector}.

\subsection{Hardware Adaptations} 
\label{sec:blackout-hardware}
Following past approaches in the League, we have added an onboard camera and a compute module for vision processing and decision-making. A Logitech C922 webcam was chosen due to its low-distortion parameters, allowing for easy camera calibration with high precision. As for additional computation, a 4GB NVIDIA Jetson Nano Developer Kit was chosen due to its small size, low power consumption, high throughput on DNN and image processing, and extensive documentation for NVIDIA libraries. Also, its System-on-Module (SoM) architecture leaves room for future improvements, by adapting our electronics to connect the module directly to our mainboard, for instance, saving even more space.

\begin{figure}[ht]
	\centering
	\captionsetup[subfloat]{justification=centering}
    \subfloat[Third floor with Jetson Nano, power supply module and onboard camera.]{\includegraphics[width=.59\linewidth, scale=1]{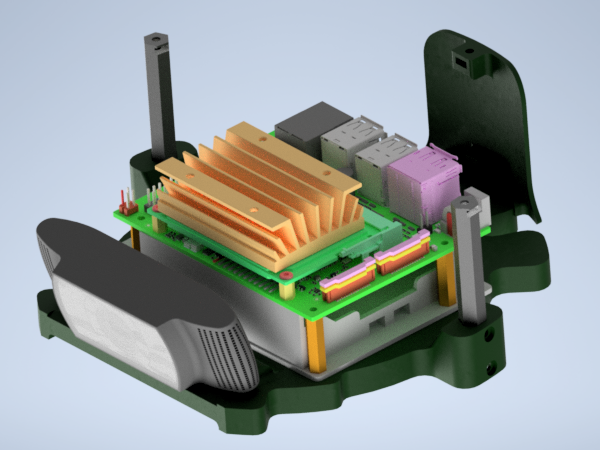}}
    \label{fig:third-floor}
    \subfloat[Vision blackout robot assembled.]{\includegraphics[width=.39\linewidth, scale=1]{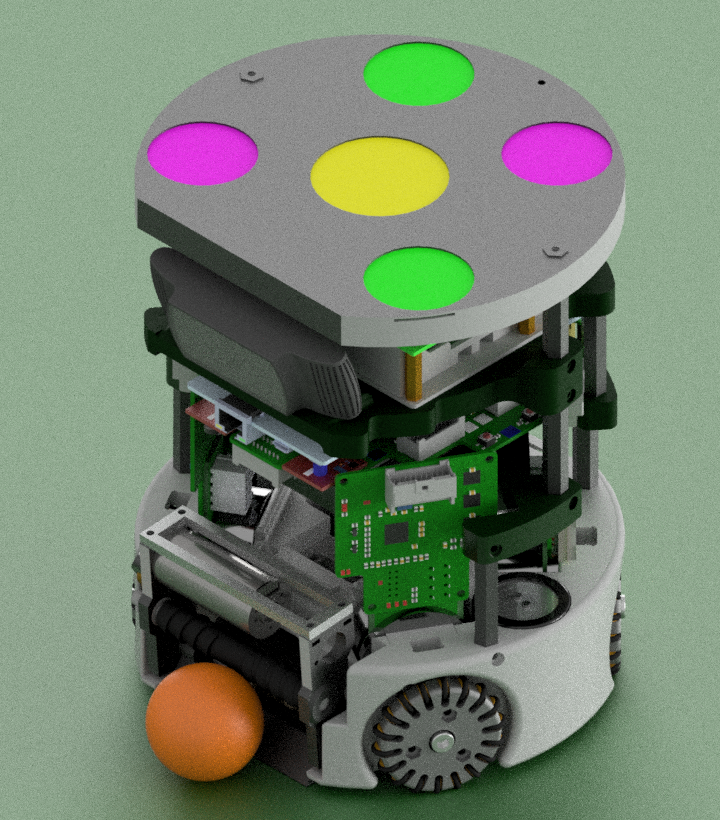}}
	\label{fig:blackout-robot-Assembled}
	\caption{Robot hardware adaptations for vision blackout challenge.}
	\label{fig:blackout-robot}
\end{figure}

Besides the Jetson Nano and the Logitech camera, we have also added a power supply module, using 4 cells of 18650 batteries, for powering this new subsystem. A new cover plate, which we call the robot's third floor, was designed for mounting those parts onto the robot and it is shown in Figure \ref{fig:blackout-robot}. It also has housings for additional standoffs, enabling us to place a SSL tag on the robot's top, which is useful for experiments and evaluation.

\subsection{Software Workflow}
The autonomous SSL robot is mainly operated by two processing modules: the Jetson Nano and the STM32F746ZI, an ARM Cortex-M7 Microcontroller Unit (MCU), also referred to as STM32F7 for simplicity. 
They communicate through an Ethernet cable using User Datagram Protocol (UDP) Socket packets.

For embedded vision, we use a Logitech C922 camera with 30 frames per second capture rate using 640x480 pixels of resolution. Vision frames are processed by the Jetson Nano running a CNN-based Object Detection model, namely SSDLite MobileNetv2\cite{ssd, mobilenetv2}, for detecting SSL objects' bounding boxes, which are used for estimating their relative positions to the robot by using pre-calibrated intrinsic and extrinsic camera parameters, as presented in \cite{objects-detection-and-position-estimation}. The paper also shares details of model retraining and deployment using TensorRT optimizations.

Decision-making is also implemented on the Jetson Nano, which runs Finite State Machines (FSMs) that implement each of the robot's autonomous skills for solving Vision Blackout challenge stages. Objects' relative positions are used as inputs and the FSM computes a target position and orientation, a navigation type, and command flags such as odometry resetting, capacitor charging, and kicking. This information is encoded into a protobuf message and sent to the MCU through the UDP connection.

At the MCU level, for our Target-Point-based navigation, we implement three movement types: Rotate-on-Self (RoS), Drive-to-Point (DtP), and Rotate-in-Point (RiP). The first accounts for rotations around the robot’s axis, mainly used for initial ball searching and self-alignment with targets. DtP implements a linear movement with orientation correction, adjusting the robot’s translation velocity according to its rotation error and distance to the target. Lastly, RiP executes a circular trajectory around a point, allowing the robot to search for a goal while looking at the ball, for instance.

\begin{figure*}[ht]
\centering
\includegraphics[width=.9\linewidth]{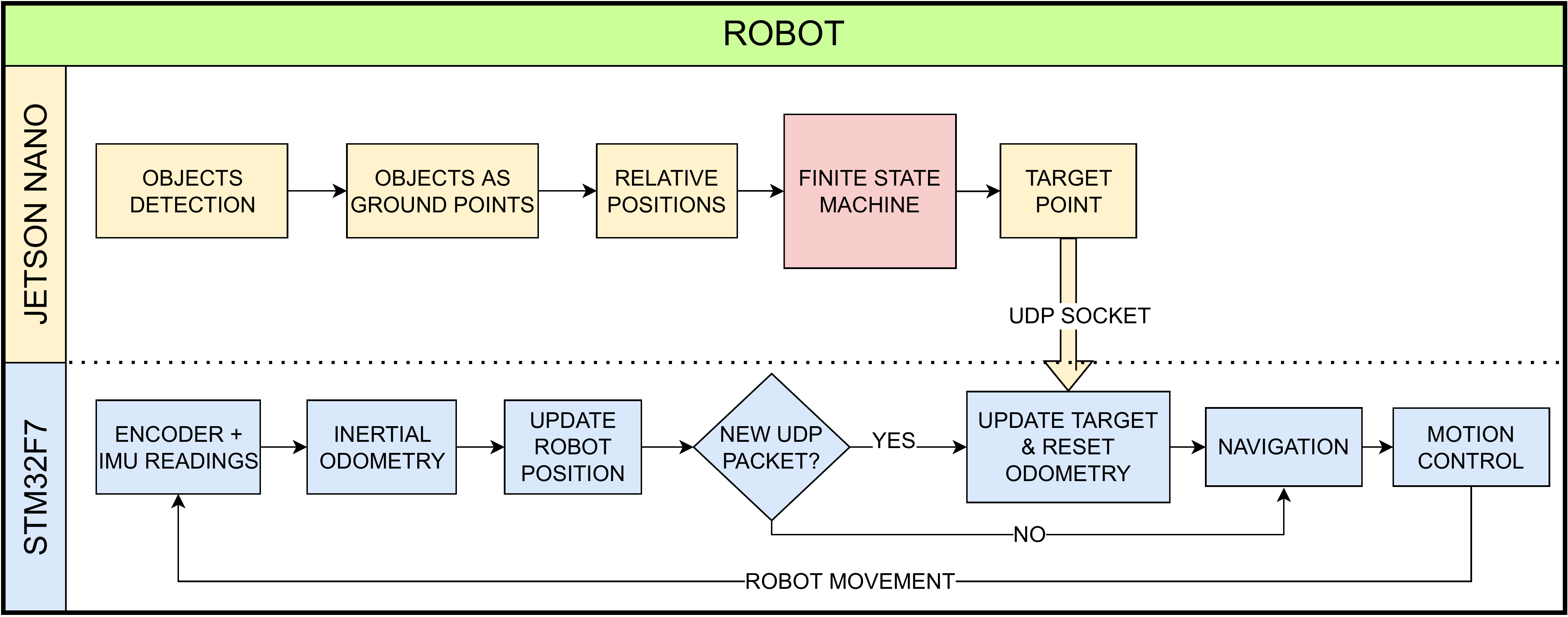}
\caption{Overview of the proposed logic diagram in order to build an autonomous RoboCup Small Size League Robot. All the modules are inside the robot, the upper modules run on the Jetson Nano, while the lower ones run on the STM32F7 MCU. Finite State Machines are defined for implementing soccer skills.}
\label{fig:overview}
\end{figure*}

The MCU also calculates the robot’s inertial odometry by computing inverse kinematics from encoder readings, and the trajectory is estimated using gyroscope measurements combined with odometry, allowing it to adjust its path while embedded vision information is not available. Figure \ref{fig:overview} illustrates an overview of the proposed architecture, and we present more details in \cite{towards-an-autonomous-ssl-robot}.

\subsection{Robot Skills}
This architecture was employed to solve the 4 stages of Vision Blackout challenge 2022. However only stages 1 and 2 were fully completed during the competition, showing our solutions were still not robust enough and highlighting many difficulties and necessary improvements, which we discuss in the next subsection.

In more recent experiments\cite{towards-an-autonomous-ssl-robot}, for evaluating our system's capabilities and weaknesses, we have executed multiple tries on different scenarios of three common SSL tasks: grabbing a ball (I), scoring on an empty goal (II), and passing the ball (III). The same rules and scoring criteria as the 2022 Vision Blackout challenge\cite{vision-blackout-2022} were applied in the tasks, except for passing the ball, which excludes scores from the kicker robot from the challenge’s stage 4. Also, we consider that the robot has succeeded in the task if conditions for all positive scores are satisfied.

Table \ref{tab:blackout-results} shares an overview of the experiments' overall results from \cite{towards-an-autonomous-ssl-robot}, showing that the robot was able to stop with the ball touching its dribbler and score a goal in 80\% of the attempts on tasks 1 and 2. As for the third task, the ball hit the receiver robot's dribbler on 46.7\% of the 15 attempts, although the robot was hit in 80\% of them.

\begin{table}[h]
\centering
\caption{Autonomous SSL Robot's Overall Performances on Proposed Tasks.}
\label{tab:blackout-results}
\def\arraystretch{1.2}
\setlength{\tabcolsep}{1.0em}
\begin{tabular}{|c|c|c|c|c|}
\hline
\textbf{Metrics} & \textbf{Task I} & \textbf{Task II} & \textbf{Task III}\\ \hline
Min Time (s)     & 6.09            & 11.89            & 9.82             \\ \hline
Max Time (s)     & 10.27           & 60.00            & 20.00            \\ \hline
Mean Time (s)    & 7.70            & 19.01            & 14.21            \\ \hline
Success Rate     & 12/15           & 12/15            & 7/15             \\ \hline
Total Score      & 40/45           & 54/60            & 47/60            \\ \hline
Penalties        & -               & 8                & 3                \\ \hline
\end{tabular}
\end{table}

\subsection{Major Issues and Ongoing Improvements}
\subsubsection{Issues from RoboCup}
One major difficulty we faced at the 2022 Vision Blackout challenge was to detect the ball at high distances, since our object detection approach was only capable of detecting it for up to 5 meters, which led us to failures at 2 of the 3 tries on stages 1 and 2. Also, our self-localization methods were not robust enough for solving stage 3, resulting in low scores and long execution times. As for stage 4, even though the passer robot was able to detect the ball, the kicker one could not move due to communication issues, leading to 3 failures.

\subsubsection{Issues from Evaluation Experiments}
During experiments from \cite{towards-an-autonomous-ssl-robot}, which results are reported in Table \ref{tab:blackout-results}, analysis from embedded vision logs have shown that most failures were caused by false positive detections from objects outside the field, highlighting the importance of discarding out-of-field information. Also, many penalties were caused due to the robot's inability of detecting field lines, and ball searching was the most time-consuming part of the tasks.

\subsubsection{Ongoing Improvements}
For discarding out-of-field information, we have been developing field boundary detection solutions, which also enable more complex exploration strategies for objects' searching, as we can avoid leaving the field, being also a useful feature for overcoming our major issue from RoboCup: not finding the ball.

Introducing a self-localization solution for SSL robots is also a necessary improvement. It enables planning more efficient paths and avoiding penalties, such as entering the defender's area. In addition, objects' searching, which was shown to be the most time-consuming part of the tasks, could be optimized using localization knowledge for more efficient field exploration. Thus, we are working on a Monte Carlo Localization (MCL) algorithm that fuses our inertial odometry approach with vision information from detected goals and field boundaries relative positions for regressing the robot's pose over time, 
following motion and observations models, and resampling techniques from typical approaches of other RoboCup leagues \cite{monte-carlo-localization1}. 

\section{Software} \label{unification}

SSL-Coach was our first version of stable software developed for the SSL category. It has a modular architecture, inspired by the STP (Skill, Tactics, and Plays)\cite{browning2005stp}, combined with the team's previous experience in the software development for robot soccer in the VSSS (IEEE Very Small Size Soccer) category. However, the resulted software presented both tightly coupled information processing and software decision-making stages, besides having a confusing data flow, due to the variety of demands and the short development period for the team's first participation in RoboCup, in 2019.

Over the years of SSL-Coach development, the accumulation of technical debts in the architecture and development infrastructure made it difficult to make improvements, such as creating more elaborate and collaborative plays among the robots, as explored in STP. These complications in software evolution are reflected in the difficulty of including new information flows and functionalities that require additional changes in other parts of the code, including critical parts of the flow that can bring side effects. Also, it has been noted as difficult to integrate new team members, transfer knowledge, and renew the team using the existing architecture. Thus, we decided to concentrate efforts on building a new code base, aiming to reduce software coupling and complexity in order to reduce execution errors, which led to failure points in the initial software architecture.

RobôCIn participates in other robot soccer categories that require the development of specialized software, namely 2D Simulation and VSSS. These categories needed to solve similar problems and situations, and as introduced in TDP 2022\cite{tdp:rc-2022}, they use similar technologies based on C++ allied to the Qt Framework to create User Interfaces (UI). However, they had a great distance from the code base, which led to duplication of effort or replication of logic to develop, and low interchangeability of developers between categories.

After surveying the RobôCIn categories' technical debts to be solved, we decided to model a more flexible and modern architecture, bringing different possibilities for expansion and reuse.Soccer-common was developed as an open-source library, used as a submodule, aiming to concentrate the global part of common code of the different team modalities, and providing a separation between UI and back-end. Also, it has as its main components: a library of geometric functions, a graphical interface with drawing support at any point, debugs, and the design of a module.

A module, in the new architecture, consists of the major abstraction capable of executing logic in parallel, providing support for communication with the visualization interface and parameters, and communicating with other modules. Also, a module can be indexed for occasions where it is intended to have multiple identical execution steps, such as the behavior of a robot. Its conception consists of the main core of our architecture, which we will describe below.

From now on, we will called Unification, the new development base software, which has already replaced our VSS-software and SSL-Coach.

\subsection{Architecture}

The tight coupling of processing in SSL-Coach is due to how its modules were created and communicate with each other. Each module consists of a thread that executes a singleton, a static global object with a unique instance at any point, and the exchange of information between these modules is done through direct connections by setters and getters, which violates the Single-Responsibility Principle (SRP)\cite{cleanArchitecture}, and consequently makes it difficult to change the existing execution flow and call tracking performed.

To resolve the SRP violations, the execution flow was oriented to data arrival events, using callbacks, thus enabling a coordinated flow of data and the execution of each thread. To create the callbacks infrastructure we use QT Signals \& Slots\footnote{https://doc.qt.io/qt-6/signalsandslots.html}, originally intended for the communication of objects linked to the UI in the QT framework, has a robust and simple system for using callbacks in C++, which consists of implementing the Observer Pattern to facilitate communication between components of the Framework.

By combining Signals \& Slots with our projected wrapper for safe shared access, the communication of our modules was implemented as a cascaded publisher-consumer system, where emitter functions are connected in the creation of modules to receivers, confirming each module requirement. As soon as a module receives the input and is registered, the information is stored in a critical region and waits for the next execution to be effectively consumed.

With this new infrastructure, we have also improved the communication between modules using more flexible data packages, simplifying modifications and corrections. Previously, the information we shared consisted of an extensive structure where all the relevant information to be transmitted was present. The distinction of its use was under an enumerator's responsibility. The filling of this structure was not always fully completed, as not all information was relevant to a specific message, which pollutes and makes it difficult to understand. We solved the problem using a variant type, introduced in C++17, which consists of a type-safe union, capable of aggregating different structures into a single type, enumerating each one of them. It allows us to direct the processing of a message through pattern matching, which simplifies the use of previously needed conditionals.

\begin{figure}[ht]
\centering
\includegraphics[width=1\textwidth]{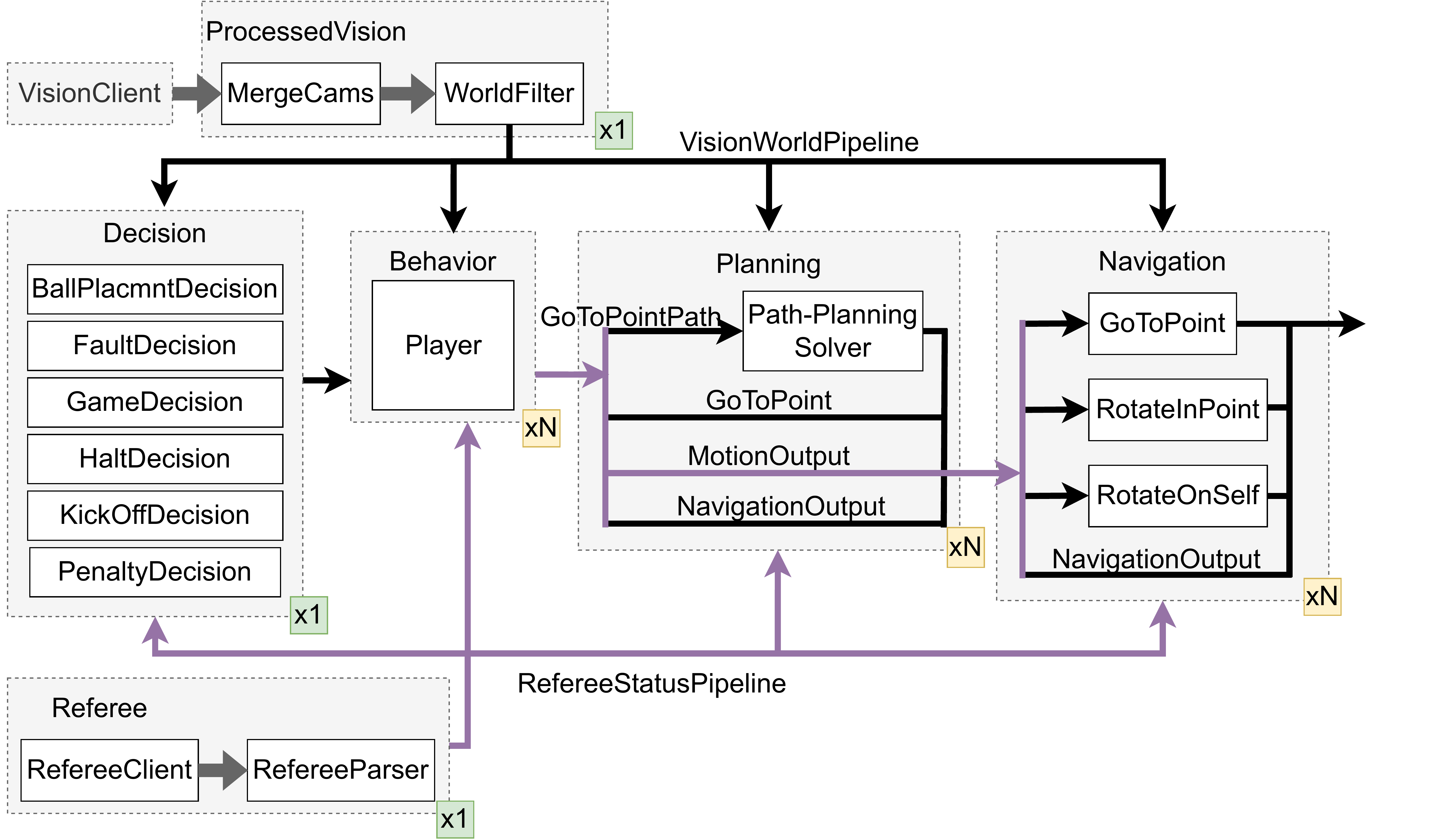}
\caption{Overview of SSL-Unification software architecture detailed dataflow. In purple is the data with variant groups type, in each module has notation of number of thread, in green modules with one thread running and in yellow modules with execution in function of number of available robots, N.}
\label{fig:overviewUnificationArchitecture}
\end{figure}

In this software architecture, the flow of information starts with the vision system that sends position information. Simultaneously, the referee sends stage and command updates. The software then receives these inputs and applies filtering processes. The decision module determines the players' behaviors (such as goalkeeper, forward, defender, and others) according to the received referee data, making decisions and using the vision data to identify which player should perform a particular action. The behavior modules then use the decision assignments to execute the intended behavior for each player in separate threads. These threads produce tactics that are processed by the planning and/or navigation modules threads in sequence. Finally, the navigation module processes the necessary actions for the robots to move, with briefed type and parameters. As a result, we achieved the architecture in Figure \ref{fig:overviewUnificationArchitecture}.

\subsection{Implementation}

We seek to reformulate modules introduced in the 2019 TDP according to the listed technical debts, reducing the number of flags and removing boilerplate code. We will describe the changes made to our code flow below:

\subsubsection{DataWorld}

In SSL-Coach, this component was responsible for receiving vision information obtained from simulators or vision software, performing vision processing and receiving commands from the referee's software. We decided to separate the processing carried out into three modules, each dedicated to one of the respective activities described above.

For the module dedicated to receiving arbitration software commands, we have developed our parser\footnote{https://github.com/robocin/soccer-common/wiki/Referee-Parser} based on the Stage and Command received, allied to the analysis of internal flags, information from the vision and the previous context, aiming at the specialization of game situations, simplifying the strategy carried out later, so each leaf situation at parser tree output, started to be treated in isolation.

\begin{figure}[ht]
\centering
\includegraphics[width=1\textwidth]{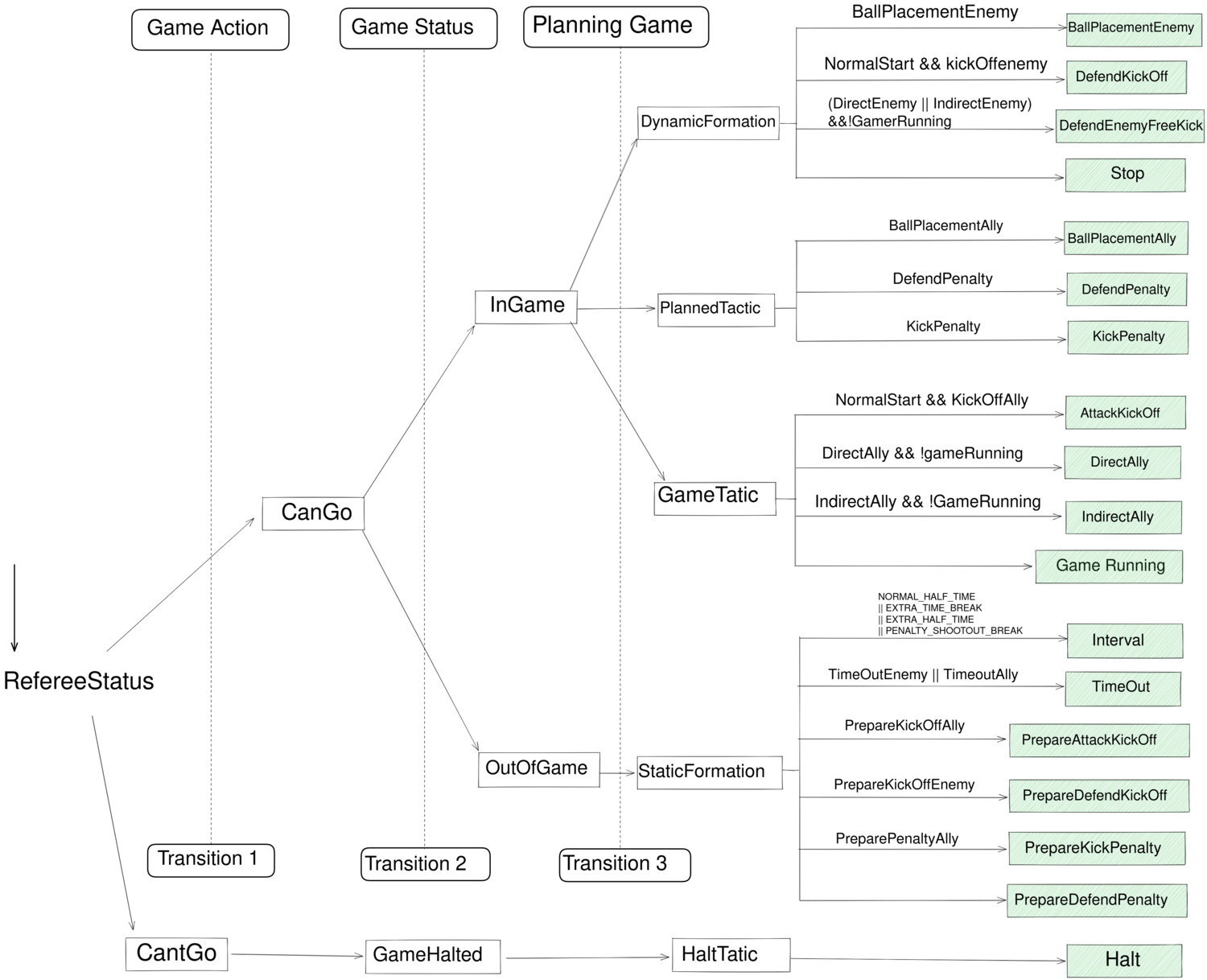}
\caption{Complete referee parser tree, showing all possible game states.}
\label{fig:refParserDiagram}
\end{figure}

The complete referee parser tree, shown in Figure \ref{fig:refParserDiagram}, starts from the game's command and the state received from the external referee. At the Game Action division, it decides if the robots must halt, or not. Then, the Game Status transition defines if we are dealing with an in-game situation or a positioning one, such as a preparation for kick-off. Lastly, at the Planning Game division, the parser chooses between states whether the robots must move without touching the ball (Dynamic Formation), execute a predefined play (Planned Tactic), or play the game normally (Game Tactic).

\subsubsection{Trainer}

With the split of the DataWorld component, and the creation of a dedicated module to parse the information received from the arbitration software, the processed commands allowed a strong restructuring in this module, where we currently make quick specific changes for opponents depending on the applied game state.

Over the years we have greatly evolved players' allocation within the decision component, the Trainer, formerly called Decision, starting from a static team with 3 Defenders, 1 Support, and 1 Forward in 2019, to a dynamic allocation based on the position of the ball and risk offered by the enemies' positions. In this way, we are currently able to be an adaptable team, but one that seeks to be extremely offensive by pressing the game to the enemy half, exchanging passes until they get an opportunity to shoot on goal.

Our offensive tactics are made up of a Forward, who is the player in possession or in a direct dispute over the ball, and of a variable number of supporters according to the position of the opposing team's robots, where each supporter seeks to stay in optimal positions, the best within our heuristics to receive a pass from Forward and perform a successful play. Once the ball is passed to the supporter, this will become the player in possession or direct dispute for the ball, switching positions: the supporter receiving the pass will become the Forward, which keeps the attack cycle performed by our team.

\subsubsection{Behavior}

With Unification, one of the team's main goals for the old Player module was to decouple functionalities and simplify state machines. In SSL-Coach, we had behaviors with finite state machines (FSM) of many states and with similar logic functions done in several different ways within the behavior itself, which made it difficult to understand the transitions to debug and make corrections.

Previously, as described, each state corresponded to an enumerator, and the state processing nodes, functions, which are incapable of storing contexts, switched by a large number of conditionals. The input for each processing node consisted of a pair $<state, context>$, with the information needed for all existing states, which made it particularly difficult to distinguish information relating only to specific states. With the architecture update, some ways to improve the implementation of an FSM for the desired purposes were also studied. Similar to the messages used for communication between modules, the machine states started to consist of a variant, while the processing nodes of these states became classes. With this change, the state processing nodes now have greater independence, with contexts capable of restarting as a transition to a new state is performed.

Also, in the current architecture, we started to apply the concept of SkillBook coming from the STP\cite{browning2005stp}, and to define the attacker as a set of tactics that involve interacting with the ball (be it kicking to goal, giving a pass or take a penalty) that was previously all together in a single FSM, making it complex and disproportionately large to deal with various situations.

In order to facilitate maintenance and future improvements, it was decided to extract the previously existing Planning and Navigation modules within the Behavior component, thus enabling the alternation of algorithms used, as we will explain below.

\subsection{Path-Planning} 

One of the changes made to the architecture was the creation of a dedicated module for path planning. After that, we became capable of exploring path optimizations and switching the used algorithm.

This year, we changed our path-planning algorithm and optimized our low-level control. Until then, we used an evolved version of the visibility graph presented on the 2019 TDP\cite{tdp:rc-2019}, due to a bunch of changes realized over the years aimed to optimize and handle corner cases. However, it has become really difficult to maintain it given the increased code complexity.

\subsubsection{Current Problems}

One of our major issues in past competitions was the high number of fouls due to crashing and robot distance to forbidden locations, as shown in Table \ref{table:gameFouls}. Due to the yellow cards arising from those fouls, we were frequently forced to play with 5 or 4 players, which reduced massively our offensive power given the reduced number of players to compose the attack. This analysis led us to optimizations on the path-planning algorithm, since the majority of those fouls were avoidable.

\begin{table}[ht]
\centering
\caption{Collision and invasion detected during matches at RoboCup 2019, 2021 and 2022.}
\label{table:gameFouls}
\begin{tabular}{|c|c|}
\hline
\textbf{Referee foul event}                          & \multicolumn{1}{l|}{\textbf{Amount}} \\ \hline
ATTACKER\_TOO\_CLOSE\_TO\_DEFENSE\_AREA & 22                                   \\ \hline
BOT\_CRASH\_UNIQUE                      & 93                                   \\ \hline
DEFENDER\_TOO\_CLOSE\_TO\_KICK\_POINT   & 69                                   \\ \hline
\end{tabular}
\end{table}

\begin{figure}[ht]
\centering
\includegraphics[width=0.55\linewidth]{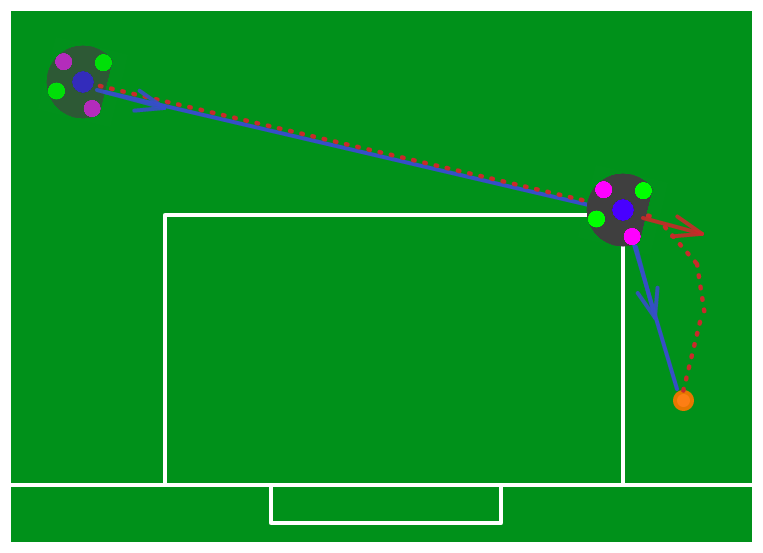}
\caption{Dissonance between the planned path in blue and the trajectory executed in red, with emphasis on the forces acting in the change of edge of the graph.}
\label{fig:robotMoveIssue}
\end{figure}

With limitations concerning Visibility Graph's nature, those related to the generated path stand out. Despite being the shortest euclidean path, it's not time-optimal for omnidirectional robots with an abrupt change in direction and velocity, as presented by Balkon et al.\cite{TimeOptimalTrajectories}. Furthermore, because the algorithm does not take into account the agent's momentum/direction, which is the whole robot's state with velocity and acceleration rather than solely position, there is a dissonance in its execution between the calculated path and the robot’s real trajectory, as shown in Figure \ref{fig:robotMoveIssue}. Moreover, the available margin for navigation error is minimal due to the generated path being tangential to the obstacles. Hence, both factors culminate in the high amount of collision and invasion, since the expansion of obstacles' boundaries is not a direct guarantee of decreased collisions in general, besides being a solution that greatly hurts our team performance.

Also, bigger obstacles reinforce some limitations with our implementation, given that as they are solely a set of points connected in the scene's graph, then it is not possible to increase the complexity of the polygons used as obstacles further than triangles and rectangles without harming the execution time. Likewise, we are not able to properly handle the escape from an obstacle, whether it is the start or target position. So, with a more generic solution, we are susceptible to dealing with a lot of corner cases, which result in both bad placements for ball disputes and defense area invasion.

\subsubsection{Desired Key Improvements}

Therefore, we listed the following sought improvements for the new algorithm:
\begin{itemize}
    \item[$\blacksquare$] Fewer number of collisions, allowing better velocity and movement.
    \item[$\blacksquare$] A generated path harmonic to the real robot's trajectory.
    \item[$\blacksquare$] Robust algorithm for real-time and dynamic scenarios as those of SSL.
    \item[$\blacksquare$] Possibility of an obstacle model that appraises the movement's dynamic, considering time as a factor to determine possible collisions.
    \item[$\blacksquare$] Obstacles that can be differentiated from each other for a greater fidelity of representation of the world.
    \item[$\blacksquare$] Possibility of simulating the robot's movement to feed estimate the robot reaching range.
\end{itemize}

\subsubsection{Solutions Adopted}

Unlike the analyzed options in TDP 2019\cite{tdp:rc-2019}, this time, we chose to study and compare Sampling and Trajectory based algorithms, which are classes of Path Planning Algorithms that had proven some robustness in terms of Motion Planning for SSL. Despite the popularity of RRT-based algorithms (RRT*, RRT-Connect, ERRT) among SSL teams, we opted for the Bang Bang trajectory\cite{bangBangTrajetory} given that its traditional implementation already suffices all of our requirements, which has a strong integration between pure planning with a series of points and the proper navigation, since it computes the robot's action velocity along the path.

Bang Bang trajectory-based path-plannings were studied and adopted by reference teams, being reported as important for the achieved results by Tigers\cite{tdp:tiger19, tigers:2019} and Er-Force\cite{tdp:er-force20, er-force:2022}. Both implementations are open-source and demonstrate distinct ways of dealing with the implementation of the algorithm. While Tigers bet on an approach based on selecting intermediate points from a constellation of points around a given origin connected by trajectory segments to the target, the Er-Force goes with a more open search approach seeking through the trajectory time and orientation.

Each implementation has its advantages and drawbacks, and we sought to validate both approaches and their code bases. We converted Tigers' implementation to C++, but the achieved performance and some discrepant behavior to the java version made us adopt Er-Force's base, which was already developed in C++ and had an execution time lower than 1ms. To achieve this execution time, the algorithm has a reduced number of search iterations and a large search bias around previously found solutions.  Thus in situations where the previous path is no longer possible and/or a large deviation is required to reach target, the search for paths fails in create trajectory. Thus, we merged some of our ideas into the algorithm, such as an additional validation of robot movement reset to prevent the speed in one direction from distorting the trajectory in its direction too much, also, ideas as well as some from the Tigers' solution, the use of the constellation of points around the robot to force further exploration of obstacle contour directions, at the cost of increased time complexity, but still contained in a time frame.

\subsubsection{Correlation with Navigation}

Path planning and navigation are inherently related. A path planning that does not considers the robot's dynamics causes a big discrepancy in the obtained result. Mainly in the Visibility Graph, which has sharp changes in the velocity and direction on the generated path. Then, by only using the $\Delta S$, the navigation needs to predict the output for the robot to fulfill and generate all the movement's state transitions that affect the planning result, but none of this feedback is propagated into the next path planning.

Aiming to close this control loop, trajectory-based algorithms are fed with the robot's current state with its position and velocity. But the software relies on data from the vision system where this current state corresponds to a robot's past state that was captured by the currently received frame. Furthermore, an SSL Robot is capable of changing its velocity in such a way that it is difficult for vision processing to keep up, therefore, approaches based only in vision limit the robot state transition update by the camera frame rate, typically 60 Hz.

The limitation of detection of the robot current state by the vision mainly impacts the ability to control the acceleration and deceleration of the robot when the path is being adjusted throughout the cycles since it cannot reach the expected state. Seeking to mitigate this effect, we developed methods for estimating the current state of the robot based on the vision frame current information, the vision processing delay and the speed commands sent to the robot during this delay, starting from the state seen in the vision, and we apply the commands sent to the robot by replicating their performance time, thus estimating its real current state. Another more effective approach to this problem to eliminate assumptions about the current state of the robot would be to calculate its current trajectory segment itself, performing embedded navigation, a solution adopted by the Tigers team that we intend to invest in the following years.

\section{Acknowledgement}

First, we would like to thank our advisors and the Centro de Informática (CIn) - UFPE for all the support and knowledge during these years of project and development. We also would like to thank all our sponsors: CESAR, Microsoft, Veroli, HSBS, Moura, and Mathworks.

%
%
%
\bibliographystyle{splncs04}
%
\bibliography{references}

\begin{thebibliography}{10}
\providecommand{\url}[1]{\texttt{#1}}
\providecommand{\urlprefix}{URL }
\providecommand{\doi}[1]{https://doi.org/#1}

\bibitem{TimeOptimalTrajectories}
Balkcom, D.J., Kavathekar, P.A., Mason, M.T.: Time-optimal trajectories for an
  omni-directional vehicle. The International Journal of Robotics Research
  \textbf{25}(10),  985--999 (2006). \doi{10.1177/0278364906069166}

\bibitem{browning2005stp}
Browning, B., Bruce, J., Bowling, M., Veloso, M.: Stp: Skills, tactics, and
  plays for multi-robot control in adversarial environments. Proceedings of the
  Institution of Mechanical Engineers, Part I: Journal of Systems and Control
  Engineering  \textbf{219}(1),  33--52 (2005)

\bibitem{dataset-ssl}
Fernandes, R., Rodrigues, W.M., Barros, E.: Dataset and benchmarking of
  real-time embedded object detection for robocup ssl. In: Alami, R., Biswas,
  J., Cakmak, M., Obst, O. (eds.) RoboCup 2021: Robot World Cup XXIV. pp.
  53--64. Springer International Publishing, Cham (2022)

\bibitem{ssd}
Liu, W., Anguelov, D., Erhan, D., Szegedy, C., Reed, S.E., Fu, C., Berg, A.C.:
  {SSD:} single shot multibox detector. CoRR  \textbf{abs/1512.02325} (2015),
  \url{http://arxiv.org/abs/1512.02325}

\bibitem{cleanArchitecture}
Martin, R.C.: Clean Architecture: A Craftsman's Guide to Software Structure and
  Design. Prentice Hall Press, USA, 1st edn. (2017)

\bibitem{objects-detection-and-position-estimation}
Melo, J.G., Barros, E.: An embedded monocular vision approach for ground-aware
  objects detection and position estimation  (2022).
  \doi{10.48550/ARXIV.2207.09851}, \url{https://arxiv.org/abs/2207.09851}

\bibitem{towards-an-autonomous-ssl-robot}
Melo, J.G., Martins, F., Cavalcanti, L., Fernandes, R., Araújo, V., Joaquim,
  R., Monteiro, J.G., Barros, E.: Towards an autonomous robocup small size
  league robot. In: 2022 Latin American Robotics Symposium (LARS), 2022
  Brazilian Symposium on Robotics (SBR), and 2022 Workshop on Robotics in
  Education (WRE). pp.~1--6 (2022).
  \doi{10.1109/LARS/SBR/WRE56824.2022.9996004}

\bibitem{tdp:tiger19}
Ommer, N., Ryll, A., Geiger, M.: Tigers mannheim (team interacting and game
  evolving robots) extended team description for robocup 2019 (2019), roboCup
  Small Size League, Mannheim, Germany, 2019

\bibitem{bangBangTrajetory}
Purwin, O., D'Andrea, R.: Trajectory generation for four wheeled
  omnidirectional vehicles. In: Proceedings of the 2005, American Control
  Conference, 2005. pp. 4979--4984 vol. 7 (2005).
  \doi{10.1109/ACC.2005.1470795}

\bibitem{vision-blackout-2022}
RoboCup: {RoboCup 2022 SSL Vision Blackout} technical challenge rules,
  \url{https://robocup-ssl.github.io/technical-challenge-rules/2022-ssl-vision-blackout-rules.pdf}

\bibitem{er-force:2022}
Robotics Erlangen e.V. Team: Open Source Framework (2022),
  \url{https://github.com/robotics-erlangen/framework}

\bibitem{monte-carlo-localization1}
R{\"o}fer, T., J{\"u}ngel, M.: Fast and robust edge-based localization in the
  sony four-legged robot league. In: Polani, D., Browning, B., Bonarini, A.,
  Yoshida, K. (eds.) RoboCup 2003: Robot Soccer World Cup VII. pp. 262--273.
  Springer Berlin Heidelberg, Berlin, Heidelberg (2004)

\bibitem{mobilenetv2}
Sandler, M., Howard, A.G., Zhu, M., Zhmoginov, A., Chen, L.: Inverted residuals
  and linear bottlenecks: Mobile networks for classification, detection and
  segmentation. CoRR  \textbf{abs/1801.04381} (2018),
  \url{http://arxiv.org/abs/1801.04381}

\bibitem{tdp:rc-2022}
Silva, C., Alves, C., Silva, E., Martins, F., Cavalcanti, L., Maciel, L.,
  Vinícius, M., Monteiro, J.G., Moura, J.P., Cruz, J.V., Santana, P.H., Sousa,
  R., Rodrigues, R., Fernandes, R., Morais, R., Araújo, V., Silva, W., Barros,
  E.: Robôcin extended team description paper for robocup 2022 (2022), robocup
  Small Size League, Recife, Brazil, 2022

\bibitem{tdp:rc-2019}
Silva, C., Martins, F., Machado, J.G., Cavalcanti, L., Sousa, R., Fernandes,
  R., Araújo, V., Silva, V., Barros, E., Bassani, H.F., de~Mattos~Neto,
  P.S.G., Ren, T.I.: Robôcin 2019 team description paper (2019), robocup Small
  Size League, Recife, Brazil, 2019

\bibitem{tigers:2019}
Tigers Mannheim Team: Open Source Software and Hardware (2019),
  \url{https://tigers-mannheim.de/index.php?id=65}

\bibitem{tdp:er-force20}
Wendler, A., Heineken, T.: Er-force 2020 extended team description paper
  (2020), roboCup Small Size League, Erlangen, Germany, 2020

\end{thebibliography}

\end{document}